\documentclass[11pt]{article}
\usepackage{acl2016}
\usepackage{times}
\usepackage{url}
\usepackage{latexsym}
\usepackage{color}

\usepackage{amsmath}
\usepackage{amssymb}
\usepackage{amsfonts}
\usepackage{tabu}
\usepackage{xcolor}
\usepackage{graphicx}
\usepackage{bm}
\usepackage{multirow}

\DeclareMathOperator*{\softmax}{softmax}

\aclfinalcopy % Uncomment this line for the final submission
\title{Unnamed Entity Recognition of Sense Mentions}

\author{Ndapa Nakashole \\
	Computer Science and Engineering  \\
	University of California, San Diego\\
	La Jolla, CA 92093 \\
	{\tt nnakashole@eng.ucsd.edu} \\}

\date{}

\begin{document}
\maketitle
\begin{abstract}
We consider the problem of recognizing mentions of human senses in text.
% an under-explored aspect of common sense knowledge acquisition. 
Our contribution is  a  method for acquiring labeled data, and a learning method that is trained on this data.
Experiments show the effectiveness of  our proposed data labeling approach and our learning model on the task of   sense recognition in text.

\end{abstract}

\section{Introduction}

Information extraction methods produce structured data in the form of  knowledge bases of factual assertions. Such knowledge bases are useful for supporting inference, question answering, and reasoning \cite{Bollacker2008,hoffartetal2012a,MitchellCHTBCMG15}. However, progress on the common sense front, as opposed to named entities such as locations, and people,  is still  limited \cite{havasi2007conceptnet,DBLP:conf/aaai/TandonMW11}. 
%Even  crowd-sourced common sense knowledge bases, spanning decades of manual effort,  for example ConceptNet~\cite{havasi2007conceptnet}, are still sparsely  populated.
% %as we show in our experiment
In this paper, we study entity recognition of common sense concepts. Our goal is to detect  mentions of concepts that are discernible by sense. For example, recognize that  ``chirping birds" is a mention of an audible concept (sound), and    ``burning rubber"  is a  mention of an olfactible concept (smell). 
We aim to detect \textit{mentions} of  concepts without  performing
co-reference resolution  or clustering  mentions. Therefore,
our setting resembles the established task of entity recognition~\cite{DBLP:conf/acl/FinkelGM05,DBLP:conf/conll/RatinovR09}, with the difference being that we focus on un-named entities.
% where some of the features useful for named entity recognition do not apply.

%\end{enumerate}
\noindent \textbf{Contribution.}
One of the factors impeding progress in common sense information extraction is  the lack of training data. It is relatively easy  to obtain labeled data for named entities such as  companies and  people. Examples of  such  named entities  can be found in  structured forms on the Web, such as  HTML lists and tables, and  Wikipedia infoboxes \cite{DBLP:conf/www/WuW08,DBLP:conf/icdm/WangC08}.  
This is not the case for common sense concepts.
We therefore propose a data labeling method, that leverages  crowd-sourcing and large corpora.  This approach provides the flexibility to control the size and accuracy of the available  labeled data for model training.  Additionally, we propose and train several sequence models including variations of  recurrent neural networks that learn to recognize mentions of sound and smell concepts in text. In our experiments, we show that the combination of our mixture  labeling approach, and  a suitable learning model are an effective solution  to sense recognition in text.

\begin{figure}[t]
	\centering
	\includegraphics[width=1\columnwidth]{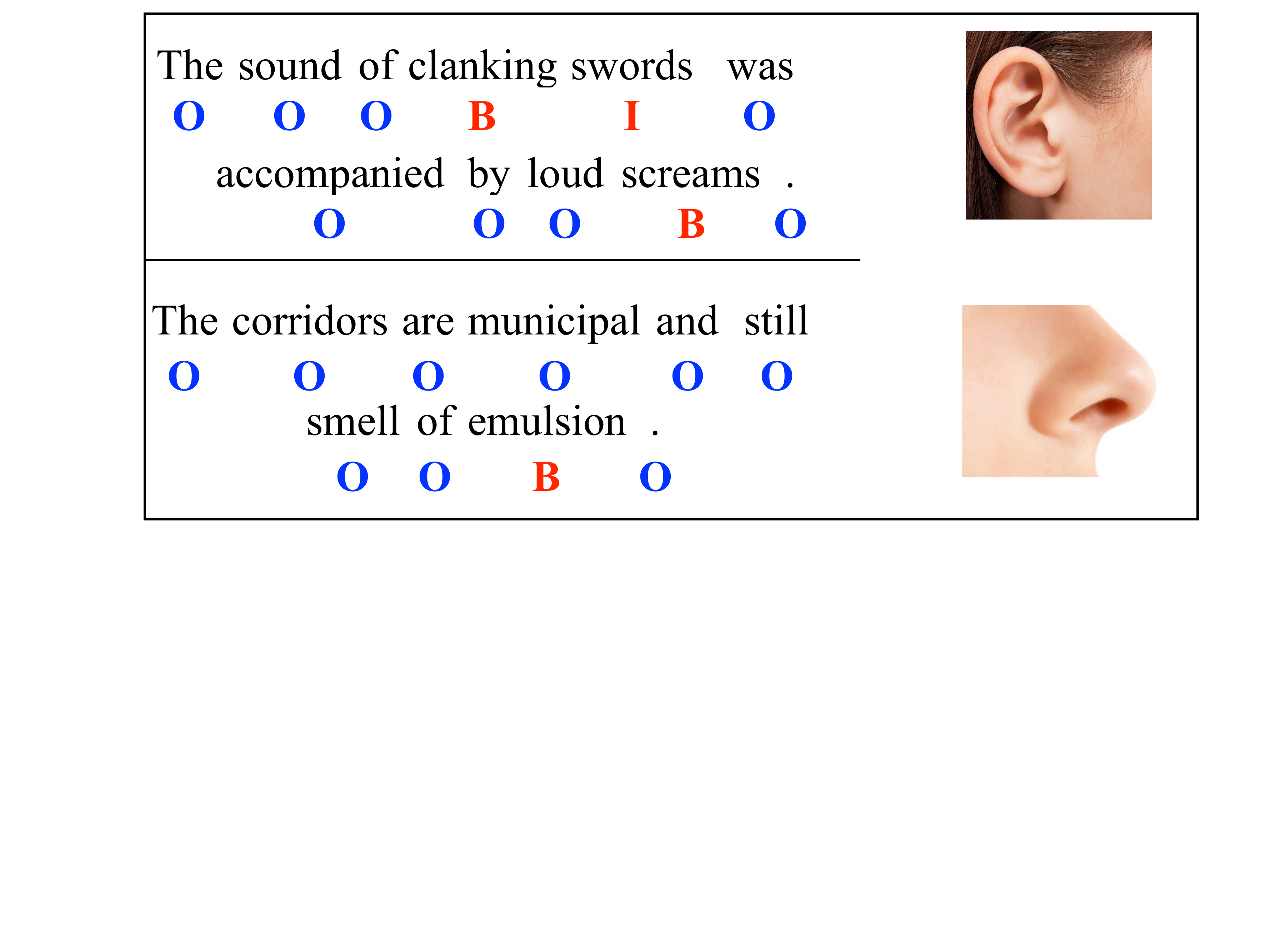}
	\vspace{-3.2 cm}
	\caption{Example  beginning-inside-outside (BIO) labeled  sentences  with mentions of  sound (top) and smell (bottom) concepts.   }
	\label{labeledsentence}
\end{figure}
\section{Problem Definition}
We would like to detect mentions of concepts discernible by sense. In this paper, we focus on  mentions of \textit{audible (sound) }and \textit{olfactible (smell) }concepts. We treat sense recognition in text as a sequence labeling task where each sentence is a sequence of tokens labeled using the  BIO tagging scheme \cite{DBLP:conf/conll/RatinovR09}. The BIO labels denote  tokens at the \textit{beginning, inside, and  outside} of a  relevant mention, respectively.  Example BIO tagged sentences are shown in Figure \ref{labeledsentence}.

\section{Data Labeling Methodologies}
There is a lack of easy to identify labeled data on the Web for common sense  information extraction,  an issue which affects named-entity centric information extraction to a  lesser degree \cite{DBLP:conf/icdm/WangC08,DBLP:conf/www/WuW08}.
We consider three data labeling approaches:
% to generating the labeled data required by our models for recognizing audible and olfactible concepts: 
\textit{i}) Automatically generate training data using judiciously specified patterns.\textit{ ii)} Solicit input on crowd-sourcing platforms. \textit{iii)} Leverage both i) and ii) in order to overcome their respective limitations.

\subsection{Pattern-based Corpus  Labeling}\label{sec:patterngenerated}
To label data with patterns,  we begin by specifying  patterns that we apply to a large corpus.
For our concepts of interest, sound, and smell, we specify the following two patterns. 
%We begin with  patterns: 
%\begin{eqnarray*}
``\text{ sound} \text{ of } \text{ \textless y\textgreater}", and 
``\text{ smell} \text{ of } \text{ \textless y\textgreater}", 
%\text{ image} \text{ of } \text{ \textless Y\textgreater}.
%\end{eqnarray*}
We then apply these patterns to a large corpus.
 In our experiments, we used the English part of  ClueWeb09. \footnote{http://lemurproject.org/clueweb09/}. The result   is a large  collection of  occurrences such as:
%\begin{eqnarray*}
``\text{ sound} \text{ of\textit{ breaking glass}}",
``\text{smell} \text{ of  \textit{perfume}}", \text{etc.}
%\end{eqnarray*}
The collections contains 134,473 sound phrases, and 18,183 smell phrases.

\begin{figure}[t]
	%\centering
	\vspace{-0.5cm}
	\includegraphics[width=1\columnwidth]{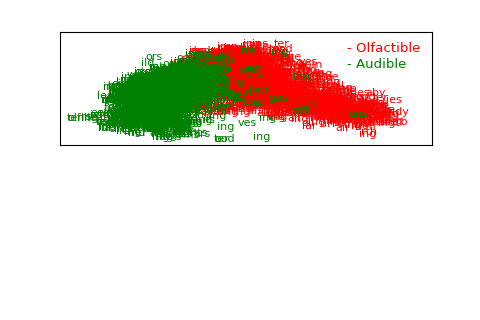}
	\vspace{-3.5cm}
	\caption{A PCA projection of  the embeddings of  audible and olfactible phrases labeled  by the pattern approach.}
		% of section \ref{sec:patterngenerated}. For readability, each data point is shown as the last three characters of  the concept name.}
	\label{fig:soundnsmell}
\end{figure}

% For qualitative analysis of the pattern-generated data, we produced 
 Figure~\ref{fig:soundnsmell}, shows  a 2D projection of the 300-dimensional  word vectors\footnote{https://code.google.com/archive/p/word2vec/}  of the discovered audible and olfactible  phrases.  We see a strong  hint of two clusters.
% , indicating that the patterns are not generating random noun phrases.
%    In the Figure, we see that the audible and olfactible noun phrases are  separated into two clusters. The patterns are not  generating completely random noun phrases.
  We later provide a quantitative analysis of this  data.

\subsection{Crowd-Sourced Supervision}
The second way of  obtaining labeled data that we consider is crowd-sourcing. We used the Amazon Mechanical Turk crowd-sourcing platform. 
%Tasks on MTurk are small questionnaires consisting
%of a description and a set of questions.

\noindent \textbf{Crowd Task Definition.}
To obtain labeled examples, we could do a ``cold call" and ask
crowd workers to list examples of   phrases that refer to senses. However,  such an approach  requires crowd workers to  think of examples without clues or memory triggers.  This   is time consuming and  error prone.   Additionally,  the  monetary cost to we have to pay to the crowd sourcing platform  could be substantial. We propose to  exploit a large corpus to obtain preliminary labeled data. This enables us to only need    crowd workers to filter the data through a series of  \textit{``yes/no/notsure"}    questions. These type of questions require little effort  from crowd workers while mitigating the amount of noisy input that one could get from open-ended,   cold call,  type of questions. 
%For a given annotation instance, we present a highlighted noun phrase within a  sentence that mentions it,  task for the turkers  is to decide if the noun phrase refers to audible concept or not.  We had a separate task for  olfactible concepts.
%To set up these tasks, we used the data generated by the patterns as described in Section~\ref{sec:patterngenerated}.
We randomly selected 1000 phrases labeled by the pattern approach as described in Section~\ref{sec:patterngenerated} to be  sound/smell phrases,  500 for each sense type.   Each  phrase  was given to 3 different workers to annotate \textit{``yes/no/notsure"}. We consider a  phrase to be a true mention of the labeled sense  if the majority of the participants chose ``yes". 
This annotation task serves two purposes: 1) to provide us with human labeled examples of sound and smell concepts ii)  to provide a quantitative evaluation of pattern generated labels. 
%Since the
%participants did not always agree, we aggregate their answers. We use
%voting to decide whether  a noun phrase is a true mention of a sound or smell concept. 

\noindent \textbf{Crowd Annotation Results.}  Table~\ref{tbl:crowdsourcelabeled} is a summary of the annotation results. First, we can see that the  accuracy of the patterns is quite high as  already hinted by Figure~\ref{fig:soundnsmell}.  Second, The inter-annotator agreement rates are moderate, but lower for  olfactible  phrases.  This is also reflected by the fact that there were around 3 times  as many ``not sure'' responses in the smell annotations  as there were in the sound annotation task (27 vs 10). Nonetheless, the output of these tasks  provide us with another option for labeled data that we can use to train our models.

\begin{table}[t]
	\centering
	\begin{tabular}{|l|| r| r|}
		\hline
		& \% Majority Yes &   $Fleiss$ $\kappa$  \\
		\hline\hline
		Audible & 73.4\% & 0.51 \\ \hline
		Olfactible &  89.6\% & 0.33  \\ \hline
	\end{tabular}
	\caption{Crowd-sourced labeling of  phrases generated  by the pattern approach of section \ref{sec:patterngenerated}.} 
		%The second column shows inter-annotator agreement rates in terms of $Fleiss$ $Kappa$ $\kappa$.
	\label{tbl:crowdsourcelabeled}
\end{table}

\subsection{Joint  Pattern \& Crowd-Sourced Labeling}\label{mixedlabels}
A third way of obtaining labeled data is to leverage both pattern-based and crowd-sourced labeling approaches. 
One  central question pertains to how  we can combine the two sources in a way that exploits the advantages of each approach while mitigating their limitations. 
%The advantage of the pattern  labeled data is that we can get a lot of it for free,  the disadvantage is that is noisier than  human labeled data. The crowd-sourced labeled data on the other hand has the disadvantage that it costs money to obtain, especially that each example has to be labeled  by multiple participants as a form of  quality-control. 
%Therefore, the crowd-sourced labeled data is often limited compared to the pattern-generated labeled data. 
We seek to start with the crowd-sourced labeled, which is small but more accurate,   and expand it with the pattern-generated labeled data, which is large but less accurate. We define  a function that determines how to expand the data.
%Let $ D^p$ be  the pattern  labeled noun phrases and   $D^c$ be the   crowd-sourcing labeled noun phrases. We wish to create a new dataset $D^{pc}$ which includes all of $D^c$, and part of $D^p$ selected by a function $f$.  Thus $D^{pc} = D^c$ $ \cup $  $f(\alpha, D^{p}, D^c)$.
%We now turn to defining  $f$.
 Let $x^{c}_{i} \in D^c$   be a crowd labeled  phrase, and  $x^{p}_{i} \in D^p$  be a pattern labeled  phrase.
Then  $x^{p}_{i}$ is added to our training labeled data $D^{pc} $ if $sim(x^{c}_{i}, x^{p}_{i})  >= \alpha$  where $sim$ is the cosine similarity between the  vector representations of the  phrases.
% $x^{c}_{i}$ and  $x^{p}_{i}$.
 For vector representations of  phrases, we use the same pre-trained Google word embeddings as those used to plot Figure~\ref{fig:soundnsmell}. For  phrases longer than one word, we use vector averaging.
%  We  tune the $\alpha$ parameter  separately 
%for audible and olfactible  phrases, using cross validation on the training data. 
\begin{figure}[t]
	%\centering
	\includegraphics[width=1\columnwidth]{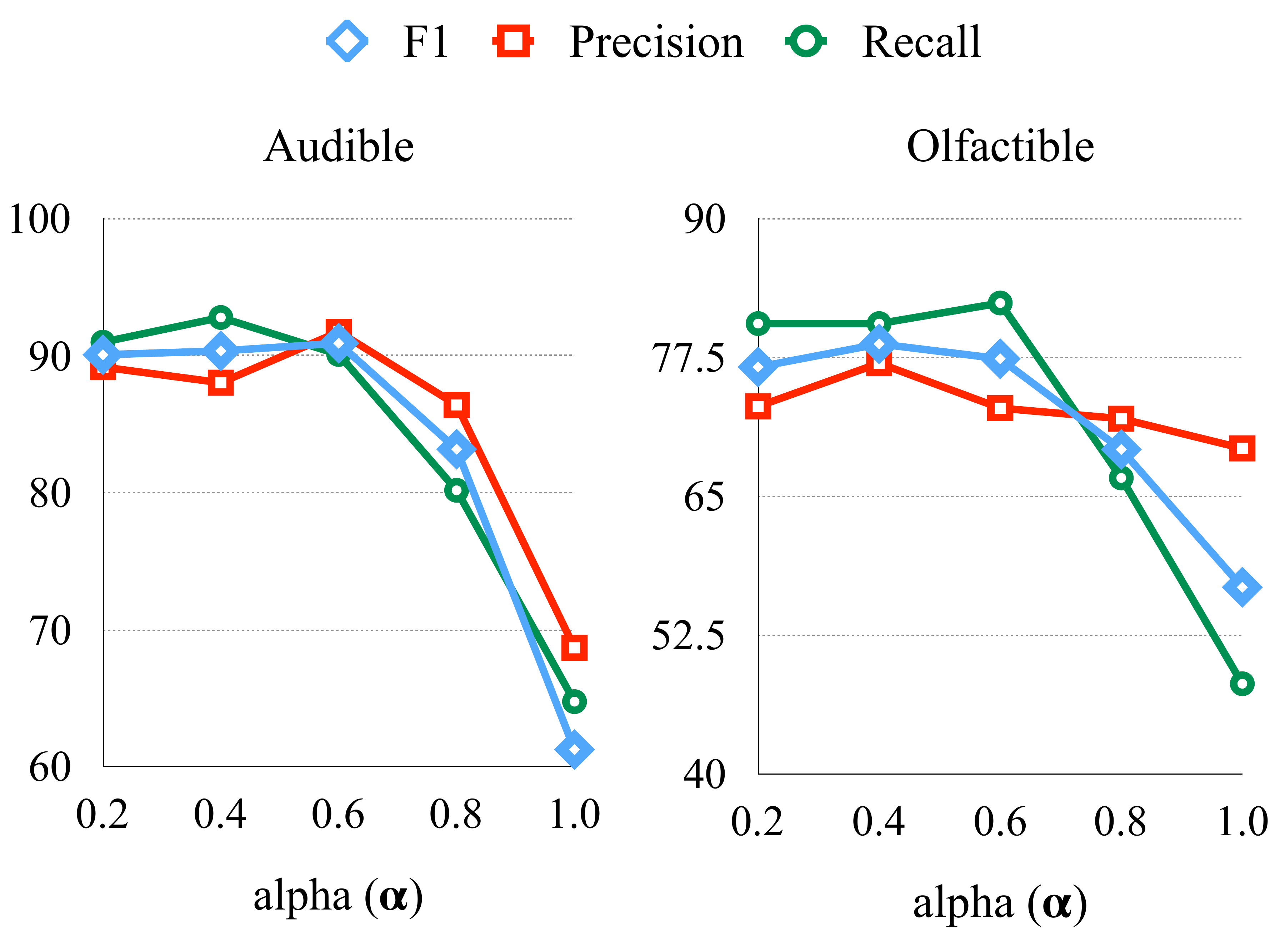}
	\vspace{-1cm}
	\caption{Performance as $\alpha$ is varied to control size and accuracy of labeled data.}
	\label{fig:alphasoundsmell}
\end{figure}
The effect of varying  $\alpha$,  for a fixed prediction model, can be seen  in Figure \ref{fig:alphasoundsmell}. When $\alpha=1$, that is we are only using the crowd-sourced labeled data, performance is at its worst. This is because even though the human labeled data is more accurate, it is much smaller, leading to potential model over-fitting problems. A more subtle finding is that  with low $\alpha$ values (i.e., \textless 0.4 for audible concepts), we have the highest recall, but not the best precision, this can be explained by the fact that, with low $\alpha$ values, we are allowing more of the automatically labeled data to be part of the training data, thereby potentially adding noise to the model. However, the advantage of the mixture approach comes from the fact that, there comes a point where precision goes up, recall slightly degrades but we obtain  the best F1 score. In Figure \ref{fig:alphasoundsmell}, we see these points at $\alpha=0.6$ and    $\alpha=0.4$ for the audible and olfactible concepts respectively. We use these values to generate the labeled data used to train models described in the rest of the paper.

%{\color{red} Major trend is that as we increase alpha, precision goes high, but we lose recall, we need to find that point when both precision and recall are high. Let us choose it.}
	\begin{figure}[t]
		%\centering
		\includegraphics[width=0.9\columnwidth]{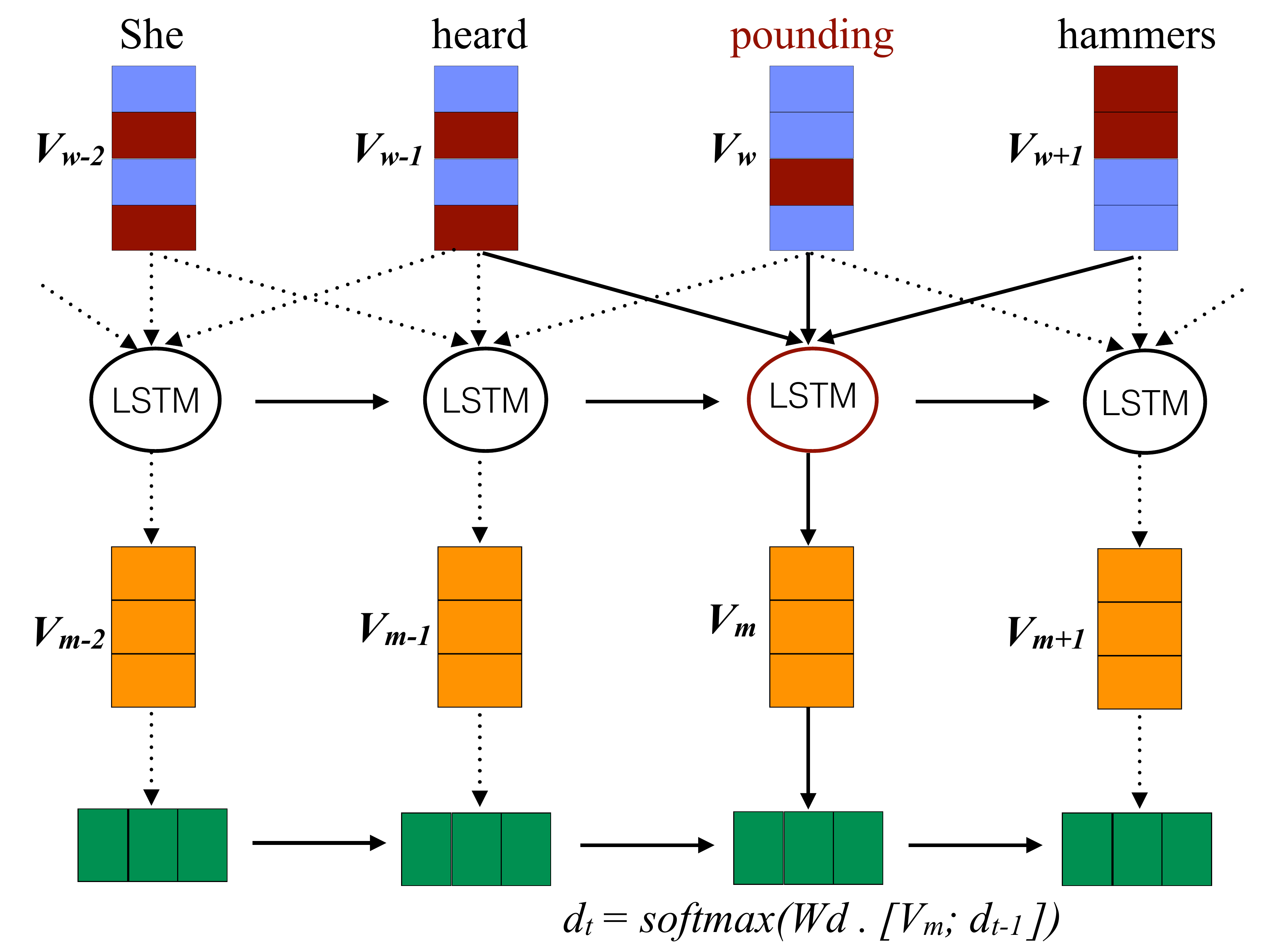}
		\vspace{-0.1cm}
		\caption{Our neural network architecture   for the task of recognizing concepts that are discernible by sensesss.}
		\label{architecture}
	\end{figure}
\section{Learning Models}
%\subsection{Sequential Prediction}
We treat sense recognition in text as sequence prediction problem, we   would like to estimate:
%\begin{equation*}
$P (y_i|x_{i-k}, ..., x_{i+k}; y_{i-l}, ..., y_{i-1})$.
%\end{equation*}
where $x$ refers to words, and $y$ refers to BIO labels.
%Where $ \bm{x}=(x_1, ..., x_n)$ where each $x_i$ corresponds to a word,  and the output sequence  $ \bm{y}=(y_1, ..., y_n)$, where each $y_i$ represents the label for the $i$-th word and where $k$ and $l$  are learned or specified depending on the choice of model.

Conditional Random Fields (CRFs) \cite{DBLP:conf/icml/LaffertyMP01}  have been widely used named entity recognition \cite{DBLP:conf/conll/RatinovR09,DBLP:conf/acl/FinkelGM05}, a task similar to our own. While the CRF models performed reasonably well on our task,  we sought to obtain improvements by proposing and training  variations of  Long  Short Memory (LSTM) recurrent neural networks \cite{DBLP:journals/neco/HochreiterS97a}.  We found our variations of LSTM sequence classifiers to do better than the CRF model, and also better than standard LSTMs.
% we  provide more details on these models in this section.

%\subsection{Model}
%{\color{blue}  Make sure to add the visibility}
%We designed   a recurrent neural  network with LSTM cells for detecting mentions of sound concepts. 
\paragraph{Word and Character Features.} As input, the LSTM neural network model takes a sentence and, as output, produces a probability distribution over the BIO tags for each word in the sentence.  To BIO tag each word in the sentence,  we use  word  features. We chose the word features to be their word embeddings.
% which we  we use are  real-valued  embeddings. 
%Each word $w$ is  mapped to a  $d$-dimensional  vector $\bm{v_w} \in \mathbb{R}^d$ through an embedding matrix $\bm{E} \in \mathbb{R}^{ |V| \times d} $, where $|V|$ is the vocabulary size, and each row corresponds to a vector of a  word.
 As additional  features, we model the character composition of words in order to capture  morphology. Neural encodings of character-level features have been shown to  yield performance gains in natural language tasks \cite{DBLP:conf/emnlp/LingDBTFAML15,DBLP:journals/tacl/ChiuN16}.   In all our experiments, we initialize the   word embeddings with the Google news  pre-trained word embeddings \footnote{https://code.google.com/p/word2vec/}. The character embeddings are learned from scratch.

\paragraph{Prediction and Output Layer Recurrence.}We represent each word as a mention within a short context window of length $m$.  We use the LSTM to encode these windows contexts in order to make a prediction for each word.  The LSTM window encoding is then used to make predictions over the BIO labels.  The output for each word is decoded by a linear layer and a  \textit{softmax} layer  into probabilities over the BIO tag labels.  Crucially, we modify the standard LSTM  by modeling  temporal dependencies by introducing a recurrence in the output layer. Therefore, the prediction  $\bm{d}_t$ at time step $t$ takes into account the prediction $\bm{d}_{t-1}$ at the previous time $t$-$1$.  Formally, we have:
%\begin{equation}\label{predwithcontext}
$\bm{d}_t = \softmax(\bm{W}_d \cdot [ \bm{v}_m ; \bm{v}_{c_a}; \bm{v}_s ; \bm{d}_{t-1}] )$, 
%\end{equation}		
where 	$\softmax(z_i) = e^{z_i}/\sum_{j}e^{z_j}$.  We  illustrate the model in Figure \ref{architecture}.
 We found this model to consistently perform  well on the  senses of sound and smell.

\paragraph{Model Evaluation.}
	\begin {table}[t]
	\centering
	\begin{tabular}{|l|l|}
		\hline
		\textbf{Sound }& \textbf{Smell} \\
		\hline
		honking cars  & burning rubber \\
		snoring & chlorine  \\
		gunshots&  citrus  blossoms \\
		live music &  fresh paint \\
		\hline
	\end{tabular}
	\label{tbl:soundsmellexamples}
	\caption{Examples of  sound  and smell concepts recognized by our method.}
	\end {table}

%\begin {table}[t]
%\centering
%\begin{tabu}{|[1.2pt]l|[1.2pt]l|p{1.6cm}|[1.2pt]}
%	\tabucline[1.2pt]{-}
%	&  \multicolumn{2}{c|[1.2pt]}{\textbf{Examples}} \\
%	& Audible  &  Olfactible \\
%	\cline{2-3}
% \textbf{Pattern}&    X= sound & X = smell\\
%	\tabucline[1.2pt]{-}
% X   of (DT) VBG NN(S)  &screeching tires & burning rubber\\
% \cline{1-3}
%X of VBG	&  snoring  &  n/a \\
% \cline{1-3}
% X of (DT) NN(S) VBG & bells ringing &  something burning  \\
%  \cline{1-3}
% X of (DT) NN(S) &  gunshots  &  chlorine \\
%  \cline{1-3}
%  X of (DT) NN NN(S) &   fire alarms &  citrus \newline blossoms \\
%   \cline{1-3}
% X of (DT) JJ NN(S) &  live music &  fresh paint  \\
%
%	%& \textbf{Total} & 134,473 \\
%		\tabucline[1.2pt]{-}
%\end{tabu}
%\caption{The part of speech generalized patterns. Examples instances are provided for
%	X=sound.  }
%\label{patterns}
%\end {table}

% What is training. what is test data.
To evaluate the models, we set aside $200$ of the $1000$ crowd-annotated phrases  as test data, meaning we have $100$ test instances  for each sense type (sound/smell). The rest of the data, $400$ per sense type was used for generating  training data using the combined crowd and pattern approach described in Section~\ref{mixedlabels}.  We set $\alpha=0.6$ and    $\alpha=0.4$ , based on Figure \ref{fig:alphasoundsmell},  for audible and olfactible concepts respectively. With these $\alpha$ values, the combination approach produced \textit{1,962} and \textit{1,702} training instances for audible and olfactible concepts respectively

% for  It is worth noting  that we ensured that the sentences in the test data do not occur in the training or validation data. We use the metrics of  F1, recall, and precision as measured by the widely used  CONLL  BIO encoding evaluation script \footnote{http://www.cnts.ua.ac.be/conll2000/chunking/conlleval.txt}\\

Performance of the various models is shown in Table~\ref{tbl:performance}. The abbreviations denote the following: \textbf{LSTM} refers to a vanilla LSTM model, using only word embeddings as features,  \textbf{+ OR} refers to the LSTM plus the output recurrence, \textbf{+ CHAR} refers to the LSTM plus  the character embeddings as features. \textbf{+ OR + CHAR} refers to the LSTM plus  the output recurrence and  character  embeddings as features. For the \textbf{CRF}, we use the commonly used features for  named entity recognition: words, prefix/suffices, and part-of-speech tag \cite{DBLP:conf/conll/RatinovR09}.
%(1) current word $x_i$,  (2) word type: is-title, (3) word type: all-digits, (4) word type:  all-uppercase. (5) prefixes and suffices of $x_i$ (6) words in the window $c= (x_{i -1}, x_i, x_{i+1})$, (7) part of speech tags for words in the window $c$. 
We can see that for both senses,  the model that uses both character embedding features, and an output recurrence  layer yields the best  F1 score. Examples of sounds and smells our method can recognize are shown in Table~\ref{tbl:soundsmellexamples}.

\begin {table}[t]
\centering
\begin{tabular}{| l|l|l|l|}
	\cline{1-4}
	\textbf{Method} & F1 & P  &  R\\
	\cline{2-4}
	\tabucline[0.5pt]{-}
	& \multicolumn{3}{c|}{\textbf{Audible}} \\
	CRF & 89.38 &  87.83 & 90.99 \\
	LSTM & 89.64 & 88.87 & 90.42 \\
	+ OR 	&  89.780 &  88.60 &  90.99 \\
	+CHAR  & 87.78 & 88.18 &  87.39 \\
	+ OR + CHAR & \textbf{90.91} & 91.740 & 90.09 \\
	%+ OR + ATTEND & 90.55 & 90.68 &  90.42 \\
	\tabucline[0.5pt]{-}
	& \multicolumn{3}{c|}{\textbf{Olfactible}} \\
	CRF & 75.73 &  79.59 &  72.22 \\
	LSTM	&   69.96  &  62.96 &  78.70 \\
	+ OR  	&  78.380 &  76.320 & 80.56 \\
	+ CHAR  	&  69.57 &  60.69 & 81.48 \\
	+ OR + CHAR & \textbf{78.73} &  76.990 & 80.56 \\
	%+ OR +  ATTEND &  79.63 & 84.70 &  75.13 \\
	\tabucline[0.5pt]{-}
	\hline
\end{tabular}
\label{tbl:performance}
\caption{Performance  of the various models on the task of sense recognition.}
\end {table}
\section{Related Work}
 Our task is related to entity recognition however in this paper we focused on  novel types of entities, which can be used to improve extraction of common sense knowledge.  Entity recognition systems are traditionally based on a sequential model, for example a CRF, and   involve feature engineering \cite{DBLP:conf/icml/LaffertyMP01,DBLP:conf/conll/RatinovR09}. 
 More recently, neural approaches have been used for named entity recognition \cite{DBLP:conf/conll/Hammerton03,DBLP:journals/jmlr/CollobertWBKKK11,DBLP:journals/corr/SantosG15,DBLP:journals/tacl/ChiuN16,shimaoka2016attentive}.  Like other neural approaches, our approach does not require feature engineering, the only features we use are  word and character embeddings. Related to our proposed recurrence in the output layer is the work of  \cite{DBLP:conf/naacl/LampleBSKD16} which introduced a CRF on top of LSTM for the task of named entity recognition. 
 
\section{Conclusions}
We have presented a method for recognizing 
concepts that are discernible by sense.
% Up to now, this particular type  of common sense knowledge
%has only been  sparsely covered in existing  knowledge bases, if at all. 
The concepts our method recognizes present opportunities for discovering additional types
of common sense knowledge, for example, learning relationships that encode information such as   which objects produce which sounds, in which environments can certain sounds be found, what is the sentiment of various types of smell, etc. These type of relations   can significantly improve coverage of common sense in knowledge bases, thereby improving their utility.

\clearpage
\nocite{*}
\bibliographystyle{acl2016}
\bibliography{sound}

\end{document}